\documentclass{article}

\usepackage{microtype}
\usepackage{graphicx}
\usepackage{subfigure}
\usepackage{booktabs} 

\usepackage[utf8]{inputenc} 
\usepackage[T1]{fontenc}    
\usepackage{url}            
\usepackage{booktabs}       
\usepackage{amsfonts}       
\usepackage{nicefrac}       
\usepackage{microtype}      
\usepackage{xcolor}         
\usepackage{tabularx}
\usepackage{subcaption}
\usepackage{multicol}
\usepackage{multirow}
\usepackage{tabularray}
\usepackage{siunitx}
\usepackage{colortbl}
\usepackage{graphicx}
\usepackage{lipsum}
\usepackage{nicefrac}
\usepackage{tabularray}

\usepackage{algorithm}%
\usepackage{algorithmicx}%
\usepackage{algpseudocode}

\usepackage{pifont}
\usepackage{marvosym}
\usepackage{fontawesome}
\usepackage{enumitem}
\usepackage{wrapfig}

\usepackage{pgfplots,gincltex}
\usepgfplotslibrary{groupplots}
\pgfplotsset{compat=1.6}
\usepackage{tikz}
\usetikzlibrary{arrows.meta}

\usepackage{hyperref}

\usepackage[accepted]{icml2025}

\usepackage{amsmath}
\usepackage{amssymb}
\usepackage{mathtools}
\usepackage{amsthm}

\usepackage[capitalize,noabbrev]{cleveref}

\theoremstyle{plain}

\theoremstyle{definition}

\theoremstyle{remark}

\usepackage[textsize=tiny]{todonotes}

\definecolor{pltblue}{RGB}{174, 199, 232}
\definecolor{pltorange}{RGB}{255, 229, 204}
\definecolor{pltgreen}{RGB}{204, 229, 204}
\definecolor{pltred}{RGB}{229, 204, 204}
\definecolor{pltpurple}{RGB}{239, 218, 230}

\definecolor{tabblue}{HTML}{1f77b4}
\definecolor{taborange}{HTML}{ff7f0e}
\definecolor{tabgreen}{HTML}{2ca02c}
\definecolor{tabred}{HTML}{d62728}
\definecolor{tabpurple}{HTML}{9467bd}

\definecolor{cblue}{RGB}{173, 201, 233}
\definecolor{clblue}{RGB}{222, 234, 246}
\definecolor{corange}{RGB}{255, 152, 67}
\definecolor{lorgange}{RGB}{255, 221, 149}
\definecolor{vividred}{RGB}{255, 0, 0} 

\definecolor{icmlblue}{rgb}{0.21,0.49,0.74}
\definecolor{yellow}{rgb}{1, 1, 0.7}
\definecolor{orange}{rgb}{1, 0.85, 0.7}
\definecolor{red}{rgb}{1, 0.7, 0.7}

\def\ie{\emph{i.e.,\ }}

\crefname{figure}{Fig.}{Figs.}
\Crefname{figure}{Fig.}{Figs.}
\crefname{table}{Tab.}{Tabs.}
\Crefname{table}{Tab.}{Tabs.}

\icmltitlerunning{}

\begin{document}

\twocolumn[
\icmltitle{CityLoc: 6DoF Pose Distributional Localization for Text Descriptions in Large-Scale Scenes with Gaussian Representation}



\icmlsetsymbol{equal}{*}

\begin{icmlauthorlist}
\icmlauthor{Qi Ma}{CVL}
\icmlauthor{Runyi Yang}{INSAIT}
\icmlauthor{Bin Ren}{INSAIT,Pisa,Trento}
\icmlauthor{Nicu Sebe}{Trento}
\icmlauthor{Ender Konukoglu}{CVL}
\icmlauthor{Luc Van Gool}{CVL,INSAIT}
\icmlauthor{Danda Pani Paudel}{INSAIT}

\icmlaffiliation{CVL}{Computer Vision Lab, ETH Zurich}
\icmlaffiliation{INSAIT}{INSAIT, Sofia University “St. Kliment Ohridski"}
\icmlaffiliation{Pisa}{University of Pisa}
\icmlaffiliation{Trento}{University of Trento}
\end{icmlauthorlist}
\icmlcorrespondingauthor{Bin Ren}{bin.ren@insait.ai}

\icmlkeywords{Machine Learning, ICML}

\vskip 0.3in
]



\printAffiliationsAndNotice{}  


\begin{abstract}
Localizing textual descriptions within large-scale 3D scenes presents inherent ambiguities, such as identifying all traffic lights in a city. Addressing this, we introduce a method to generate distributions of camera poses conditioned on textual descriptions, facilitating robust reasoning for broadly defined concepts.
Our approach employs a diffusion-based architecture to refine noisy 6DoF camera poses towards plausible locations, with conditional signals derived from pre-trained text encoders. Integration with the pretrained Vision-Language Model, CLIP, establishes a strong linkage between text descriptions and pose distributions. Enhancement of localization accuracy is achieved by rendering candidate poses using 3D Gaussian splatting, which corrects misaligned samples through visual reasoning.
We validate our method's superiority by comparing it against standard distribution estimation methods across five large-scale datasets, demonstrating consistent outperformance. Code, datasets and more information will be publicly available at our \href{https://runyiyang.github.io/projects/cityloc/}{project page}.
%
\end{abstract}
\section{Introduction}
\label{sec:intro}
\begin{figure}[t]
    \centering
    \includegraphics[width=0.96\linewidth]{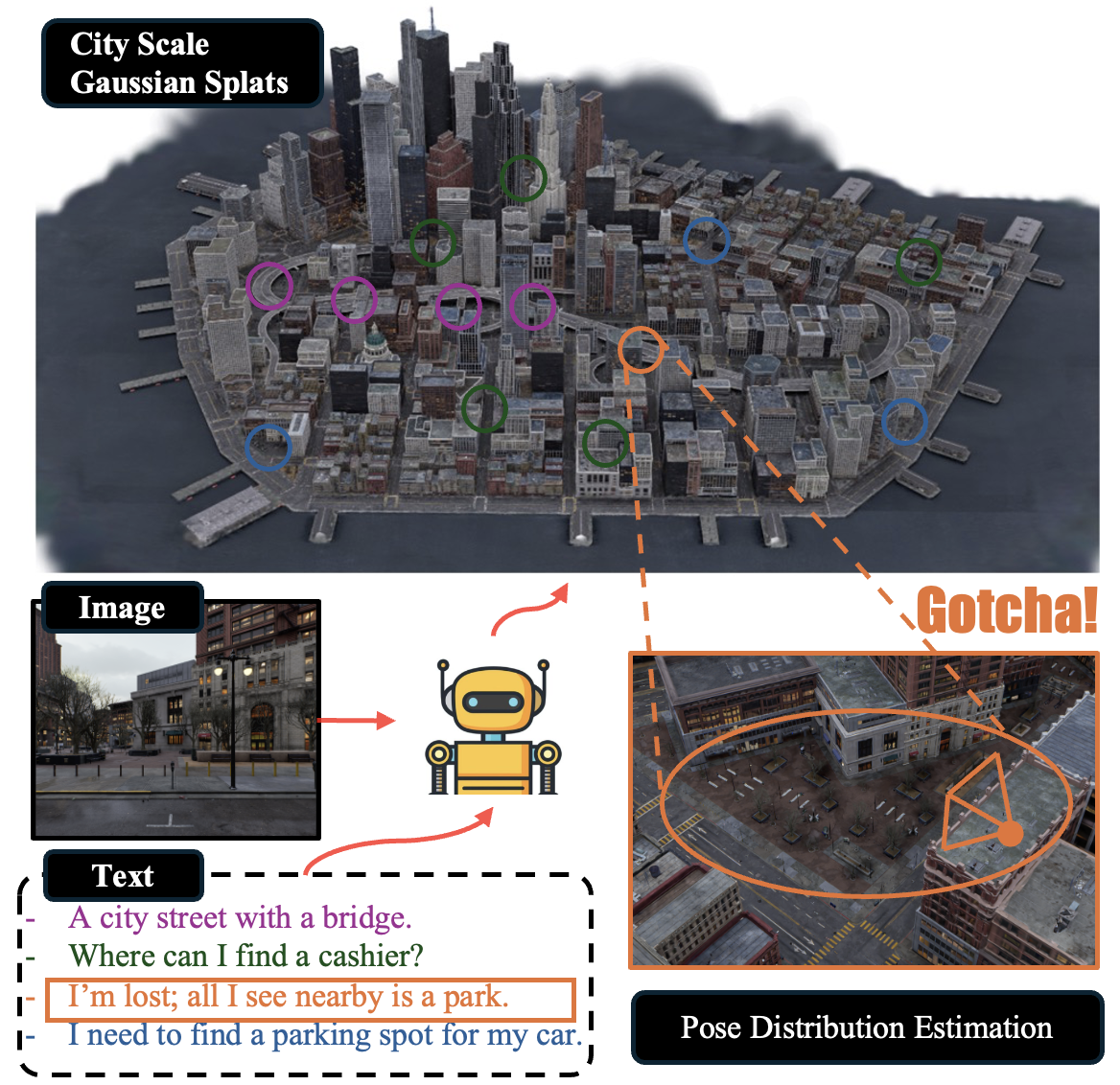}
    \vspace{-2mm}
    \caption{\textbf{CityLoc:} Given an ambiguous text description, our method accurately estimates the camera pose distribution across a large-scale urban environment, pinpointing probable locations like parking spots. Using Vision-Language Models (VLMs), our approach also incorporates image inputs for more precise, context-aware pose localization.
    }
    \label{fig:teaser}
    \vspace{-5mm}
\end{figure}
With the emergence of multi-modal understanding at scale, the integration of text and visual inputs has become increasingly intertwined. Although these modalities are complementary, they are not always available jointly. Therefore, methods such as CLIP~\cite{clip}, DALLE~\cite{DALLE}, LLAVA~\cite{llava}, and others~\cite{li2024llava, lin2024moe, wang2023visionllm, bai2023qwenvl, zhu2023minigpt4, chen2023shikra, peng2023kosmos2},  have been developed to bridge the gap between these modalities across various tasks.
In this work, we aim to estimate pose distributions within a city-scale visual environment, leveraging textual descriptions, where the 3D contextual environment is already known.
Our motivation for pursuing this research is twofold: 
(i) To decode visual locations from human language descriptions, facilitating natural human-robot interactions; 
(ii) To enable large language models (LLMs) to navigate and interpret visual scenes using language tokens. 
Our goal is to achieve text-based 6 degrees of freedom (6DoF) localization in expansive scenes where conventional feed-forward processing is impractical.

Text-based 6DoF localization in large 3D scenes, driven by our motivations, is inherently ambiguous. For instance, consider the scenario described by \emph{"building next to the street near a traffic light and zebra crossings"} which can apply to numerous locations. 
We aim to achieve distributional localization that accounts for all potential instances that fit a given textual description, \ie we intend to generate a distribution of camera poses conditioned on the text description.

To achieve this, we construct a dataset of pose-text pairs using Large Multimodal Model. We then propose a diffusion-based conditional distribution learning technique to iteratively refine noisy initial poses by conditioning on mixed visual-text features. Our approach stands in contrast to existing methods cited in ~\cite{kolmet2022text2pos,xia2024text2loc}, which seek the most probable pose and depend on highly detailed textual descriptions.

\begin{figure}[t]
    \centering
    \includegraphics[width=0.97\linewidth]{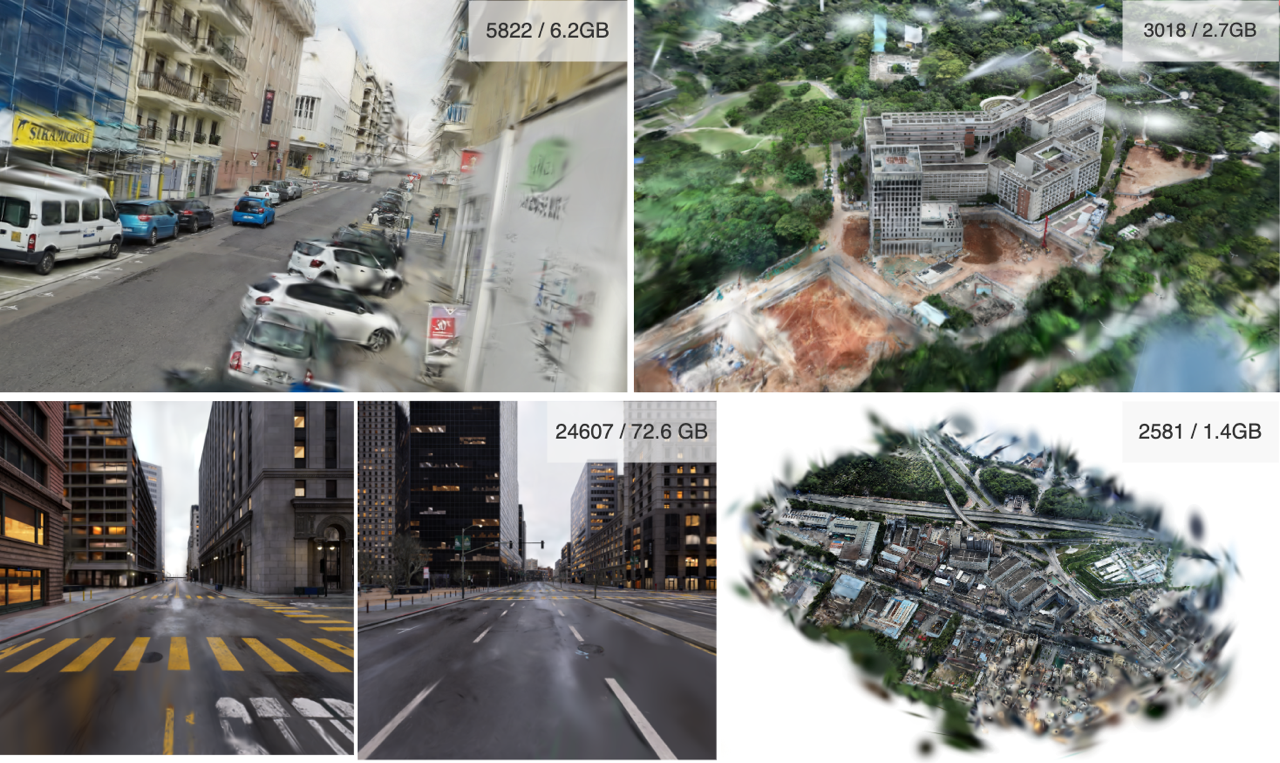}
    \caption{We present qualitative results of our large-scale Gaussian splats, including the number of images and the trained Gaussian memory size for each scene.}
    \vspace{-0.3cm}
    \label{fig:gs:quality}
\end{figure}

For visual representation, we use multiple large-scale 3D Gaussian scenes as shown in \cref{fig:gs:quality}, built in-house using a recent hierarchical 3D Gaussian method~\cite{h3dgs}. This allows us to rasterize 2D images from specified 6DoF poses, facilitating visual reasoning based on the outputs of diffusion models. While the diffusion process could be conditioned on these rasterized images to enhance localization precision~\cite{wang2023posediffusion}, for efficiency in city-scale tasks, we avoid direct conditioning. Instead, we refine pose candidates through a two-step approach: 
(i) Initial filtration of pose candidates to remove unlikely poses, and  
(ii) Subsequent refinement of poses by comparing the input text with descriptions from the rasterized 2D images.
Our method’s effectiveness is demonstrated in handling both single-modality scenarios with only text and multi-modal settings that integrate images and text.


Our method significantly differs from existing works like PoseDiffusion~\cite{wang2023posediffusion}, Text2Pose~\cite{kolmet2022text2pos}, Text2Loc~\cite{xia2024text2loc}, and RET~\cite{wang2023text}. PoseDiffusion, designed for central objects with deterministic poses, struggles with text-driven ambiguity and scale, relying on multi-view adjustments that falter in large-scale settings. In contrast, Text2Pose, Text2Loc, and RET depend on precise textual descriptions for 2D localization, lacking visual reasoning and proving inefficient in our experiments due to diverse scenes, sparse text, and the challenges of estimating 6DoF poses.

We deploy our method across 5 diverse large-scale scenes, covering over 10 square kilometers and featuring various viewpoints from street-level to aerial. These scenes include both urban and suburban environments in real and simulated formats. 
To ensure varied text descriptions during training, we utilize Llava~\cite{llava}. During testing, we evaluate performance across varying levels of description granularity, incorporating both textual and visual inputs. Our results show that more detailed descriptions significantly enhance localization accuracy, affirming our method’s effectiveness in varied settings.

The contributions of this work are:
\begin{enumerate}[nosep]
    \item Experimental Setup and Benchmarking: We established a robust experimental framework to assess city-scale, text-based 6DoF pose distribution estimation, incorporating corresponding large scale Gaussian splats as 3D representation.
    
    \item Novel Approach for Text-Based 6DoF Localization: We present a novel diffusion-based approach that integrates mixup training with multimodal image-text features using CLIP. This method effectively bridges the gap between textual descriptions and accurate spatial localization, achieving robust performance in large-scale urban environments.
    
    \item Pose Refinement Technique: We employ Gaussian splatting rendering to enhance the pose accuracy, discarding mismatches and optimizing alignment by leveraging cosine similarity with text features.  This ensures the refined pose corresponds precisely to the most relevant location based on the provided text description.

    \item State-of-the-Art Results: Our approach delivers superior performance, surpassing baseline methods in both pose estimation accuracy and distribution modeling.
\end{enumerate}

\section{Related Work}
\label{sec:related}
\paragraph{Diffusion Model.} 
Diffusion models~\cite{sohl2015deep,ho2020denoising,dhariwal2021diffusion}, inspired by non-equilibrium thermodynamics, approximate complex data distributions by reversing a noise addition process through a series of diffusion steps. Originally developed for generative tasks, these models have demonstrated impressive results in image~\cite{ho2020denoising,song2020denoising,dhariwal2021diffusion,wang2022zero}, video~\cite{singer2022make,ho2022video}, and 3D point cloud generation~\cite{lyu2021conditional,zhao2024denoising,liu2023spatio,ren2024survey}, as well as in natural language~\cite{austin2021structured,li2022diffusion} and audio generation~\cite{popov2021grad}. More recently, diffusion models have been adapted for discriminative tasks, including image segmentation~\cite{amit2021segdiff,brempong2022denoising} and visual grounding. However, there has been little exploration of applying diffusion models to camera localization tasks. In this work, we leverage the iterative nature of diffusion to refine probabilistic spatial representations for accurate camera pose estimation. By modeling spatial correlations through a Gaussian representation, we progressively refine the 6 DoF camera pose, achieving high precision and robustness in large-scale scenes, thus showcasing the flexibility of diffusion models in localization and scene understanding.

\begin{figure*}[ht]
    \centering
    \includegraphics[width=1.0\linewidth]{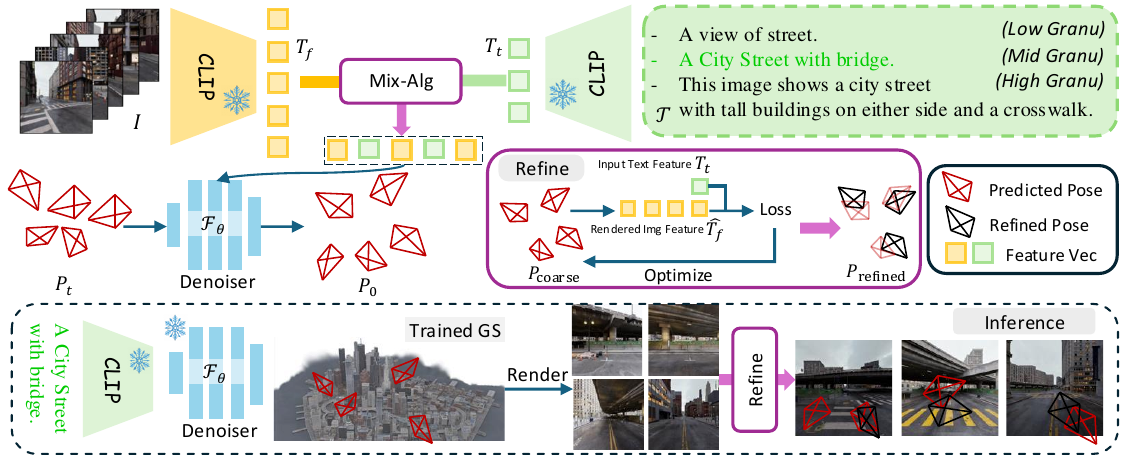}
    \vspace{-6mm}
    \caption{\textbf{Overview of CityLoc}. In the training process, where images and multi-level of granularity text input are first converted to CLIP features. A mix algorithm combines these features to train a pose diffusion model, mapping them to a 6DoF camera pose distribution. In the inference phase, where the pose diffusion model outputs camera poses for any given text input. A pretrained Gaussian representation is used to refine the poses, aligning the input text features with the rendered image features.}
    \label{fig:framework}
    \vspace{-1em}
\end{figure*}

\vspace{-0.5cm}\paragraph{Multi-Modal Large Language Models.} To extend the advancements of language models \citep{brown2020language, touvron2023llama, chowdhery2023palm, le2023bloom, hoffmann2022training} across modalities, Multi-Modal Large Language Models (MLLMs) integrate language and vision \citep{yin2023survey, liu2024visual, zhu2023minigpt, alayrac2022flamingo, zheng2024gaussiangrasper}. These models excel in tasks requiring an understanding of both text and images, which is relevant for camera localization in large-scale scenes. Flamingo \citep{alayrac2022flamingo} was among the first to align image-text pairs using gated cross-attention blocks, highlighting the importance of multi-modal integration.
In 6 DoF camera localization, MLLMs enhance the interpretation of complex urban environments by integrating semantic information from both modalities. End-to-end MLLMs often fine-tune intermediate networks \citep{lai2024lisa, zhang2023next} or sampler modules \citep{you2023ferret} to map visual features into the language space, improving scene representations for localization tasks. Models like BLIP-2 \citep{li2023blip}, MiniGPT-4 \citep{zhu2023minigpt}, and LLava \citep{liu2024visual} bridge modality gaps using querying transformers and two-stage training processes, demonstrating the effectiveness of combining visual and textual data to enhance localization accuracy.
Other notable models such as Otter \citep{li2023otter}, mPLUG-Owl \citep{ye2023mplug}, and InstructBLIP \citep{dai2024instructblip} offer architectural inspirations. Drawing from these MLLMs, our method utilizes Gaussian representations~\cite{kerbl20233d} to improve 6 DoF camera localization in large-scale scenes, enhancing both accuracy and computational efficiency.

\vspace{-0.3cm}
\paragraph{Large-scale 3D Representation and Localization.} Recent advancements in view synthesis, such as NeRFs \cite{mildenhall2021nerf, barron2021mipnerf, mueller2022instant, barron2022mipnerf360, wu2023mars} and 3DGS \cite{kerbl20233d,h3dgs,liu2024citygaussian,ma2024shapesplat,yang2024spectrally,lou2024robo}, have revolutionized 3D scene representation by utilizing differentiable rendering and optimization. To address the problem of large scenes, several methods have divided scenes into blocks or using multi-level decompositions, such as Mega-NeRF \cite{turki2022mega}, Block-NeRF \cite{tancik2022block}, CityGaussian \cite{liu2024citygaussian}, and hierarchical 3DGS \cite{h3dgs}.

Localization and mapping are closely linked, with high-quality 3D reconstructions aiding visual localization. Methods like PoseNet \cite{kendall2015posenet} predict camera poses but struggle in large environments. iNeRF \cite{yen2021inerf} and iGS \cite{sun2023icomma} invert NeRFs and 3DGS for 6DoF pose estimation. Text-based methods like Text2Pos \cite{kolmet2022text2pos} and Text2Loc \cite{xia2024text2loc} generate poses from descriptions, but can suffer from ambiguity. In contrast, our method learns a distribution of poses from text and refines them using a 3DGS map, marking the first work to bridge text descriptions with pose distributions and use Gaussian representations for large-scale pose refinement.
\begin{algorithm}[H]
    \caption{Mixup Training Algorithm}
    \label{alg:mixup}
    \begin{algorithmic}[1]
    \Require 
        \Statex \textbf{Inputs:}
        \begin{itemize}
            \item Image embedding $\mathcal{T}_f$ and text embedding $\mathcal{T}_t$
            \item Random pose $P_{\text{rand}}$ and ground truth pose $P_{\text{gt}}$
            \item Swap ratio $\beta \in [0, 1]$ (hyperparameter)
            \item Denoiser $\mathcal{F}_{\theta}$
        \end{itemize}
    \Ensure Estimate the pose distribution \( p(P \mid \mathcal{T}) \)
    \For{epoch $= 1$ to $N$}
        \For{each batch $(\mathcal{T}_f, \mathcal{T}_t, P_{gt}) \in \mathcal{D}$}
        \State Sample a timestamp $t \sim \text{Uniform}(\{1, \dots, T\})$
        \State Random Sample embeddings: 
            \[
            \mathbf{\mathcal{T}}_{\text{mix}} \gets 
            \begin{cases} 
            \mathcal{T}_t & \text{with swap ratio } \beta, \\
            \mathcal{T}_f & \text{with swap ratio } 1 - \beta.
            \end{cases}
            \]
            \State Sample noise and predict noise
            \State Predict pose: $P_{\text{pred}} \gets h_{\text{DDPM}}(\mathbf{T}_{\text{mix}}, P_{\text{rand}})$
            \State Compute loss: $\mathcal{L} \gets \|P_{\text{pred}} - P_{\text{gt}}\|^2$
            \State Update model: $\mathcal{F}_{\theta} \gets \arg\min_{h} \mathbb{E}[\mathcal{L}]$
        \EndFor
    \EndFor
    \State \Return $\mathcal{F}_{\theta}$
    \end{algorithmic}
\end{algorithm}
\section{Proposed Method}
\label{sec:method}
The CityLoc pipeline, illustrated in \cref{fig:framework}, integrates the Vision Language Model (VLM) as a condition to achieve multi-model localization. It further refines pose distribution estimation by aligning cross-modal features, minimizing the discrepancy between the rendered image feature embeddings and the input text feature embeddings. In the following sections, we will discuss each component of the pipeline in detail.

\noindent\textbf{Problem Setting.} 
We address 6-DoF pose distribution estimation by combining text and image features in a multimodal framework. This approach integrates scene descriptions (e.g., textual or visual cues) to enhance pose distribution estimation. First, we extract CLIP features for both text and image data, denoted as \( \mathcal{T}_t, \mathcal{T}_f \in \mathbb{R}^d \). Formally, given an input text feature \( \mathcal{T}_t^i \) or image feature \( \mathcal{T}_f^i \), our goal is to estimate the pose distribution \( p(P \mid \mathcal{T}^i) \) that aligns with the corresponding feature description \( \mathcal{T}^i \). This is achieved by sampling \( M \) discrete pose candidates \( \{P^{i,j}\}_{j=1}^{M} \), where each \( P^{i,j} \in \mathbb{SE}(3) \) represents a likely pose associated with the given text or image input.

\begin{algorithm}[H]
    \caption{Gaussian Refinement Algorithm}
    \label{alg:gaussian_refinement}
    \begin{algorithmic}[1]
    
    \Require 
        \Statex \textbf{Inputs:}
        \begin{itemize}
            \item Initial coarse pose $P_{\text{coarse}}$, Ground Truth pose $P_{\text{gt}}$
            \item Input text embedding $\mathcal{T}_t$ from CLIP Encoder $f_{\text{CLIP}}$
            \item City-scale Gaussian model $\mathcal{G}$
            \item Hyperparameters: learning rate $\eta$, iterations $N$
        \end{itemize}
    \Ensure Refined camera pose $P_{\text{refined}}$, success flag $\text{success}$
    
    \State Initialize $P_{\text{pred}} \gets P_{\text{coarse}}$, optimizer: $\text{Adam}(P_{\text{pred}}, \eta)$
    \State Render $\hat{I} \gets \mathcal{G}(P_{\text{pred}})$
    \State Compute $\hat{\mathcal{T}_{f}} \gets f_{\text{CLIP,img}}(\hat{I})$
    \State $\mathcal{L}_{\text{init}} \gets \frac{\hat{\mathcal{T}_{f}}^\top \cdot \mathcal{T}_t}{\|\hat{\mathcal{T}_{f}}\|_2 \cdot \|\mathcal{T}_t\|_2}$
    
    \If{$\mathcal{L}_{\text{init}} < \tau_1$}
        \State \Return $P_{\text{pred}}, \text{False}$ \Comment{Reject poor initial pose}
    \EndIf

    \For{$k = 1$ to $N$}
        \State Render $\hat{I} \gets \mathcal{G}(P_{\text{pred}})$
        \State Compute $\hat{\mathcal{T}_{f}} \gets f_{\text{CLIP,img}}(\hat{I})$
        \State $\mathcal{L}_{\text{clip}} \gets \frac{\hat{\mathcal{T}_{f}}^\top \cdot \mathcal{T}_t}{\|\hat{\mathcal{T}_{f}}\|_2 \cdot \|\mathcal{T}_t\|_2}$
        \State Backpropagate $-\mathcal{L}_{\text{clip}}$ to update $P_{\text{pred}}$
    \EndFor

    \If{$\mathcal{L}_{\text{clip}} < \tau_2$}
        \State \Return $P_{\text{pred}}, \text{False}$ \Comment{Reject refinement}
    \EndIf
    
    \State \Return $P_{\text{pred}}, \text{True}$
    \end{algorithmic}
\end{algorithm}

\subsection{Diffusion-based pose estimation} 
\label{seq:pose_via_diffusion}
CityLoc models the conditional probability distribution $p(P|\mathcal{T})$ of pose parameters $P$ given features $\mathcal{T}$.

Using a diffusion-based model \cite{sohl2015deep,wang2023posediffusion}, we model $p(P|\mathcal{T})$ with a denoising process. Specifically, $p(P|\mathcal{T})$ is estimated by training a diffusion model $\mathcal{F}_\theta$ on a training set $\mathcal{S} = \{(P_j, \mathcal{T}_j)\}_{j=1}^M$ of $M \in \mathbb{N}$ locations with input textual and visual features $\mathcal{T}_j$ and groud truth pose parameters $P_{gt}$. At inference, for a new set of text features $\mathcal{T}$, we sample $p(P|\mathcal{T})$ to estimate pose parameters $P$. The denoising process is conditioned on $\mathcal{T}$, \ie, $p_\theta(P_{t-1} \mid P_t, \mathcal{T})$:
\begin{equation}\label{eq:p_}
p_\theta(P_{t-1} | P_t, \mathcal{T}) = \mathcal{N}(P_{t-1}; \sqrt{\alpha_t} \mathcal{F}_\theta(P_t, t, \mathcal{T}), (1 - \alpha_t) \mathbb{I}).
\end{equation}
\vspace{-5mm}

\begin{figure*}[ht]
    \centering
    \includegraphics[width=0.95\linewidth]{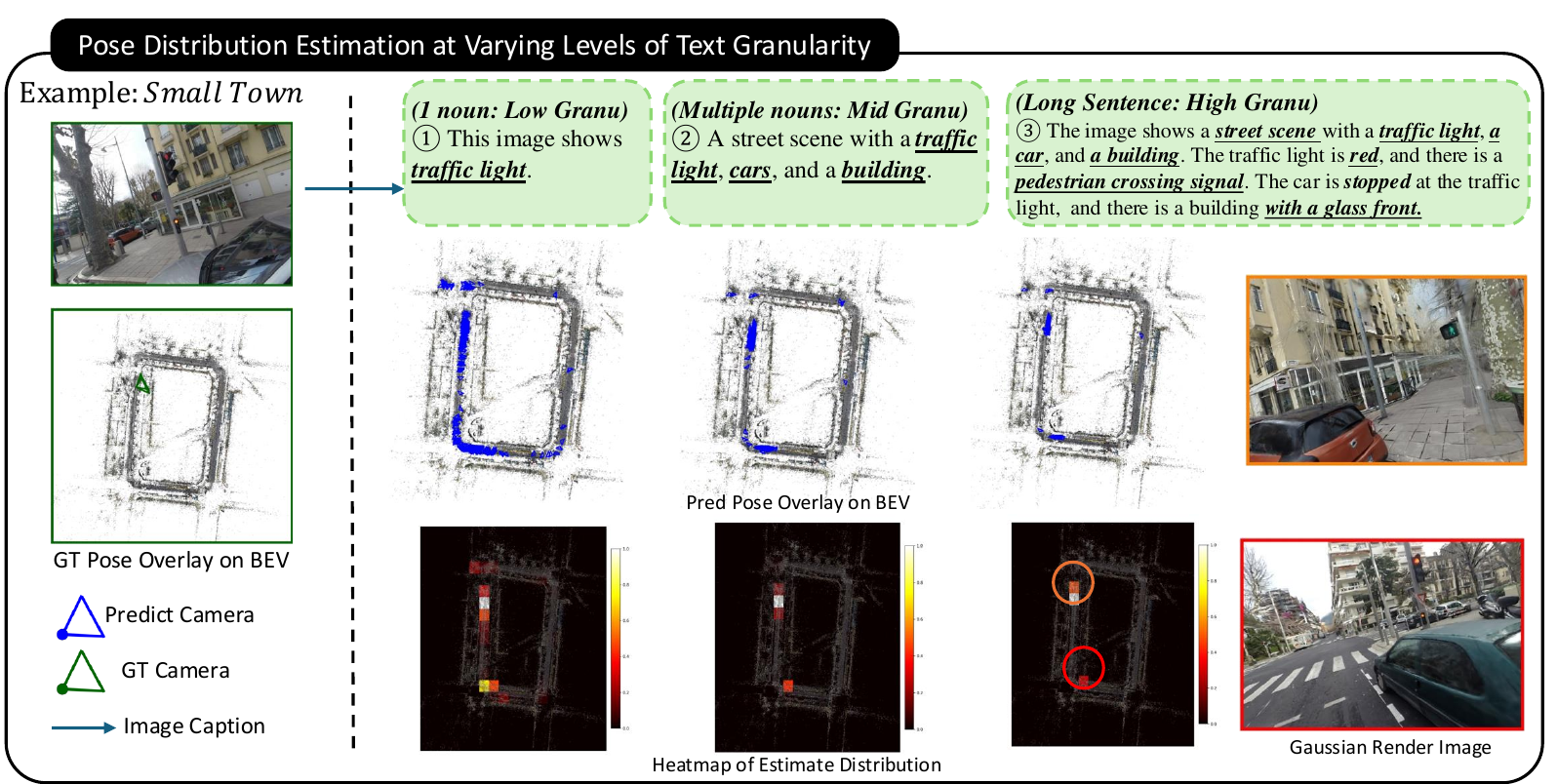}
    \vspace{-2mm}
    \caption{\textbf{Qualitative results on the small town dataset:} The enlarged \textcolor{green}{green camera} and its corresponding images represent those used to generate multiple text prompts with varying levels of granularity. We report the pose distribution conditioned on different levels of text details. The results clearly demonstrate that more informative text inputs lead to more precise location estimates. Additionally, cameras estimated in other locations provide meaningful insights. This is illustrated by selecting a pose within a high-density area for rendering as shown in \textcolor{tabred}{red camera} and \textcolor{taborange}{orange camera}, where both estimates reveal the presence of a traffic light. Zoom in for better visual results.}
    \label{fig:exp:qualitive_small_town_granu}
    \vspace{-1em}
\end{figure*}

\paragraph{Denoiser $\mathcal{F}_\theta$.}
The denoiser $\mathcal{F}_\theta$ is implemented as a Transformer:
\begin{equation}\label{eq:denoiser_transformer}
\mathcal{F}_\theta(P_t, t, \mathcal{T}) =  \text{Trans}\left[\left(\text{cat}(P_t^i, t, \psi(T^i)\right)_{i=1}^N \right] = \mu_{t-1}.
\end{equation}
Here, the Transformer processes a sequence of noisy tuples $P_t^i$, time $t$, and embeddings $\psi(T^i) \in \mathbb{R}^{D_\psi}$ of text features $T^i$, outputting the corresponding denoised parameters $\mu_{t-1} = (\mu_{t-1}^i)_{i=1}^N$. 
Training $\mathcal{F}_\theta$ is supervised by the denoising loss:
\begin{equation}\label{eq:p_superv}
\mathcal{L}_\text{diff} = E_{t \sim [1, T], P_t \sim q(P_t | P_0, \mathcal{T})} \|
    \mathcal{F}_\theta(P_t, t, \mathcal{T}) - P_0
\|^2,
\end{equation}
where the expectation aggregates over all diffusion steps $t$, the diffused samples $P_t \sim q(P_t | P_0, \mathcal{T})$, and a training set $\mathcal{S} = \{(P_{0,j}, \mathcal{T}_j)\}_{j=1}^M$ of scenes with features $\mathcal{T}_j$ and poses $P_{0,j}$. Following DDPM sampling \cite{ho2020denoising}, we initialize with random parameters $P_T \sim \mathcal{N}(\mathbf{0}, \mathbb{I})$ and, at each iteration $t \in (T, ..., 0)$, sample the next step $c_{t-1}$ as:
\begin{equation}\label{eq:ddpm_sampling}
P_{t-1} \sim \mathcal{N}(
P_{t-1} ;
\sqrt{\bar{\alpha}_{t-1}} \mathcal{F}_\theta(P_t, t, \mathcal{T}), (1-\bar{\alpha}_{t-1}) \mathbb{I}
).
\end{equation} 

\vspace{-0.2cm}

\begin{figure*}[t]
    \centering
    \includegraphics[width=0.9\linewidth]{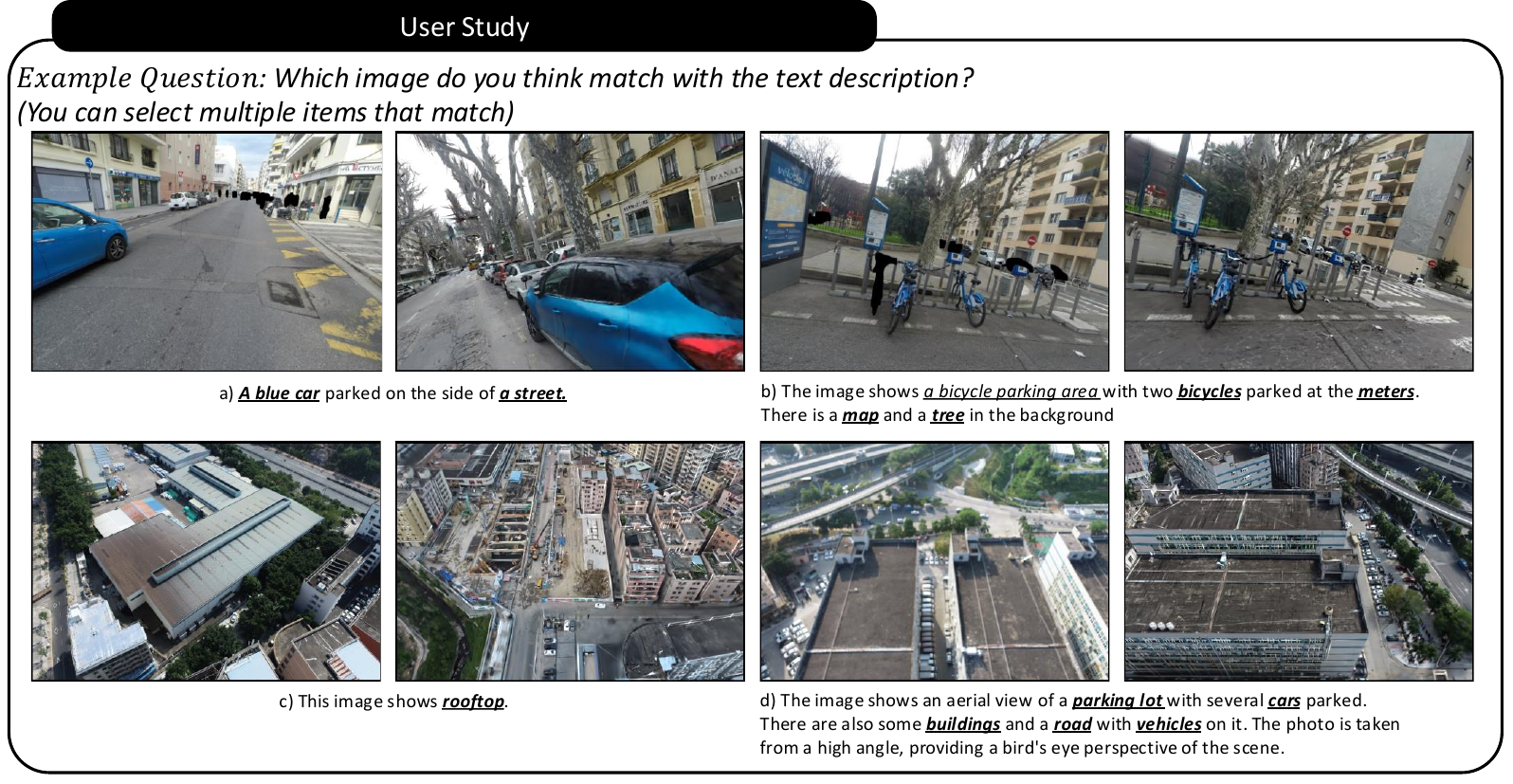}
    \vspace{-3mm}
\caption{An example question from the user study and its corresponding qualitative results. Ground-truth (GT) images appear on the left, while the rendered images are shown on the right. Quantitative results of user study please refer to \cref{tab:exp:user_study}. }
    \label{fig:exp:visual_results}
    \vspace{-1em}
\end{figure*}

\paragraph{Text-Prompt Generation.}
To construct the dataset with pose-text pairs, we extend image-pose pairs by generating captions using LLava-NEXT model~\cite{llava, liu2023llava,liu2024llavanext} to process images at varying granularities. The text generation conditions on a question argument that specifies the desired level of detail in the captions, as outlined below:
\begin{itemize}[itemsep=0pt, parsep=0pt, topsep=0pt, leftmargin=*]
    \label{template}
    \item \textbf{Nouns:} Generates a list of up to ten nouns that capture the primary visual elements present in the image. 
    \item \textbf{Template for Nouns Usage:} The extracted nouns are formatted as "A view of [nouns]" for input into the CLIP model, as seen in the upper right part of \cref{fig:framework}.
    \item \textbf{Long Sentence:} Produces a detailed caption composed of three concise sentences, providing a comprehensive description of the image. 
    \item \textbf{Mid Sentence:} Delivers a balanced description in two sentences, offering an optimal trade-off between detail and conciseness. 
    \item \textbf{Short Sentence:} Condenses the image content into a single succinct sentence, emphasizing the most prominent features. 
\end{itemize}
Afterward, we calculate embedding cosine similarity of text and image, filtering out text descriptions not matching.

\paragraph{Multi-model Conditioning}

We propose the Mixup Training Algorithm (Alg. \ref{alg:mixup}) to bridges the gap between input textual descriptions and predicted camera poses by introducing a novel embedding combination strategy during training.  First, embeddings are generated for images and texts using a CLIP Image Encoder and a CLIP Text Encoder, respectively. These embeddings are then sampled using a hyper-parameter swap ratio for each data batch shown in step 4 of (Alg. \ref{alg:mixup}). 
By minimizing the mean squared error between the predicted and ground truth poses, the model is optimized to embed multi-model information into the pose distribution estimation. This approach ensures a robsut alignment between text, image and the pose and achieves accurate pose distribution estimation in large-scale, complex environments.
By minimizing the mean squared error between the predicted and ground truth poses, the model is optimized to effectively integrate multi-modal information into the pose distribution estimation process. This optimization ensures strong generalization across diverse textual inputs by leveraging alignment between text and visual modalities through combined conditioning. As a result, the model achieves accurate and reliable pose likelihood prediction, even in large-scale, complex environments.

\subsection{Text and Gaussian based Refinement}
Unlike image-based refinement \cite{sun2023icomma}, which optimizes individual poses for a given image, our approach focuses on refining pose distributions using only text as input to improve distribution estimation. As shown in Alg. \ref{alg:gaussian_refinement}, we begin with an initial coarse pose samples $\hat{P}_{coarse}$, derived through a diffusion process conditioned on text descriptions. This estimate serves as a probabilistic initialization for our pose configuration. 

From this initial estimate, we render an image \(\hat{I_i} = \mathcal{G}(\hat{P_i})\) using rasterization from a trained Gaussian model \(\mathcal{G}\).

We compute the cosine similarity between the CLIP embedding features of the rendered image and the CLIP embedding of the text input. If the similarity score is smaller than a threshold \(\tau_1\), the sample is excluded from the distribution estimation. The negative of this similarity score is used as the loss function, which is backpropagated to refine the pose. Through this process, the coarse pose is iteratively updated to maximize the feature-level alignment with the text embedding. 
Finally, if the similarity score exceeds a second threshold \(\tau_2\), the refined sample is accepted and saved.






\section{Experiments}
\label{sec:experiments}
\subsection{Experimental Settings}
We evaluate the performance of CityLoc on a diverse set of datasets. 
First, we assess our method on the Small Town dataset~\cite{h3dgs}, a large-scale collection comprising 5,822 images covering an area of over 40,000 m\textsuperscript{2}, with camera poses extracted using COLMAP~\cite{schoenberger2016sfm}. 
Additionally, we test our approach on the UrbanScene3D dataset~\cite{UrbanScene3D}, which contains high-resolution drone imagery of expansive urban environments. The initial GPS-derived camera poses are refined through a data preprocessing procedure based on MegaNeRF~\cite{turki2022mega}. 
Finally, we evaluate CityLoc on the extensive MatrixCity dataset~\cite{li2023matrixcity}, which includes 67,000 aerial images and 452,000 street-level images spanning two city maps with a total area of 28 km\textsuperscript{2}.

\begin{table}[ht]
    \centering
    \resizebox{0.48\textwidth}{!}{%
    \begin{tabular}{l c c c c c}
        \toprule
        & \multicolumn{5}{c}{\textbf{Datasets}} \\
        \cmidrule(lr){2-6}
        & \textbf{Town} & \textbf{Residence} & \textbf{Sciart} & \textbf{Street} & \textbf{Aerial} \\
        \midrule
        \multicolumn{6}{c}{\textbf{k = 15}} \\
        \midrule
        MCDrop      & 7.72            & 4.30           & 6.29           & 1.26                     & 0.27 \\
        Ours        & 16.91           & 6.23           & 19.60          & \cellcolor{red}16.44     & 10.50 \\
        Ours Mixup  & 22.62 
                     & 10.21 
                     & 32.01 
                     & 7.96                     & 9.69 \\
        Ours Refined & \cellcolor{red}26.37 
                     & \cellcolor{red}12.42 
                     & \cellcolor{red}34.69 
                     & 15.11 
                     & \cellcolor{red}13.21 \\
        \midrule
        \multicolumn{6}{c}{\textbf{k = 10}} \\
        \midrule
        MCDrop      & 8.69           & 6.10           & 4.46           & 1.98                      & 0.25 \\
        Ours        & 15.21          & 14.01          & 28.32          & \cellcolor{red}29.86      & 19.90 \\
        Ours Mixup  & 35.33 
                     & 20.26 
                     & 43.61 
                     & 12.76                     & 18.18 \\
        Ours Refined & \cellcolor{red}41.46 
                     & \cellcolor{red}24.21 
                     & \cellcolor{red}44.39 
                     & 23.24 
                     & \cellcolor{red}20.03 \\
        \midrule
        \multicolumn{6}{c}{\textbf{k = 5}} \\
        \midrule
        MCDrop      & 3.07           & 3.20           & 0.09           & 2.45                      & 0.20 \\
        Ours        & 11.68          & 25.83          & 27.82          & \cellcolor{red}76.46      & \cellcolor{red}47.13 \\
        Ours Mixup  & 55.35 
                     & 25.93 
                     & 36.47 
                     & 27.57                     & 35.51 \\
        Ours Refined & \cellcolor{red}57.21 
                     & \cellcolor{red}28.85 
                     & \cellcolor{red}39.92 
                     & 35.21 
                     & 39.22 \\
        \bottomrule
    \end{tabular}
    } 
    \caption{RDA Performance comparison across five datasets for different methods under three values of \(k\)\ explained in \cref{fig:exp:metric}. We present three variants of our method: Ours (image only in training), Ours Mixup (image mixed with text prompt), and Ours Refined (Mixup with Gaussian splatting-based refinement). Best results per row are highlighted in \textcolor{red}{red}. Town, Street, Aerial are refer to Small Town, Matrix City Street and Matrix City Aerial respectively.}
    \label{tab:exp:test_text}
\end{table}

\begin{table}[ht]
    \centering
    \resizebox{0.48\textwidth}{!}{%
    \begin{tabular}{l c c c c c}
        \toprule
                & \multicolumn{5}{c}{\textbf{Datasets}} \\
        \cmidrule(lr){2-6}
         & \textbf{Town} & \textbf{Residence} & \textbf{Sciart} & \textbf{Street} & \textbf{Aerial} \\
         \midrule
         \multicolumn{6}{c}{\textbf{Low Granularity}} \\
         \midrule
         Ours       & 2.73  & 1.09  & 1.18   & 1.75   & 2.12  \\
         Ours Mixup & \cellcolor{red}5.96  & \cellcolor{red}10.67 & \cellcolor{red}7.97   & \cellcolor{red}1.89   & \cellcolor{red}2.72  \\
         \midrule
         \multicolumn{6}{c}{\textbf{High Granularity}} \\
         \midrule
         Ours       & 3.70  & 1.53  & 2.32   & 1.59   & 1.81  \\
         Ours Mixup & \cellcolor{red}18.69 & \cellcolor{red}24.84 & \cellcolor{red}19.89  & \cellcolor{red}2.98   & \cellcolor{red}5.43  \\
         \midrule
         \multicolumn{6}{c}{\textbf{Max Granularity}} \\
         \midrule
         Ours       & \cellcolor{red}83.75 & 35.75 & 126.70 & \cellcolor{red}105.07 & \cellcolor{red}61.70 \\
         Ours Mixup & 80.45 & \cellcolor{red}96.18 & \cellcolor{red}120.61 & 41.13  & 43.38 \\
         \bottomrule
    \end{tabular}%
    } 
    \caption{\textit{Granularity Experiments}. We compare \textbf{Ours} and \textbf{Ours Mixup} across five datasets under three granularity levels (Low, High, Max), and report the RDA value of different granularity of the methods accordingly. Town, Street, Aerial are refer to Small Town, Matrix City Street and Matrix City Aerial respectively.}
    \label{tab:exp:test_text_granu_more}
\end{table}

In our experimental setup, image description granularity varies across levels: 
for low granularity, only a list of nouns is used; 
for high granularity, all sentence types are employed for a comprehensive textual representation; 
and for maximum granularity, the noun list and all sentence types are combined to generate the most detailed image descriptions.

\begin{figure}[t]
    \centering
    \includegraphics[width=0.8\linewidth]{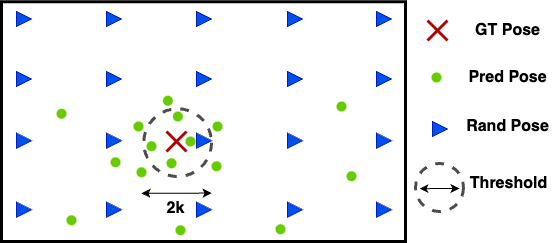}
    \vspace{-2mm}
    \caption{\textbf{Relative Distribution Accuracy (RDA)} measures the accuracy of the sample distribution within a specified region, defined by a distance k,  Translation is measured in units of 10\% of the scene scale, while rotation is measured in degrees.}
    \label{fig:exp:metric}
    \vspace{-1em}
\end{figure}

\noindent\textbf{Evaluation Metrics.} 
We report Relative Distribution Accuracy (RDA) in \cref{fig:exp:metric} to assess distribution learning performance. Unlike prior work (\cite{kolmet2022text2pos}, \cite{xia2024text2loc}), in our task, the ambiguous text input and large-scale scene make it challenging to identify all other positive samples. It is defined as the ratio of the accuracy of the predicted distribution (\(\text{Acc}_{\text{Pred}}\)) to the accuracy of a random distribution (\(\text{Acc}_{\text{rand}}\))as expressed in \cref{eq:RDA}.

\vspace{-5mm}
\begin{equation}
\text{RDA} = \frac{\text{Acc}_{\text{Pred}}}{\text{Acc}_{\text{rand}}}, \quad \text{where} \quad \text{Acc} = \frac{n_k}{N}
\label{eq:RDA}
\end{equation}
\vspace{-5mm}

\noindent\textbf{Implementation Details.} All experiments were conducted using an NVIDIA A6000 GPU. We implemented DDPM with an 8-layer transformer. For pose representation, we used a quaternion vector for rotation and a translation vector in global coordinates. The training was done with a learning rate of \(1 \times 10^{-4}\) and a CosineWarmup scheduler.
We used the Adam optimizer and trained for 30k iterations on most datasets, while for Matrix City Street, was 50k iterations.
We set $\tau_1$ as 0.17 and $\tau_2$ as 0.2 in \cref{alg:gaussian_refinement}.


\subsection{Experimental Results}
\noindent\textbf{Monte Carlo Dropout Baseline.} Monte Carlo Dropout (MCDropout) is a practical technique for approximating Bayesian inference in neural networks by applying dropout not only during training but also at inference time. By randomly deactivating a subset of neurons through dropout masks and sampling multiple forward passes, MCDropout generates an empirical distribution of model outputs. In camera relocalization, this method~\cite{kendall2016modellinguncertaintydeeplearning} estimates a distribution over predicted poses, rather than a single deterministic result. By shifting from point estimates to a broader distribution of solutions, MCDropout empowers the relocalization system to make more informed decisions and manage uncertainty in real-world applications, and also make this method a baseline for a distributional estimation of camera poses.

\noindent\textbf{Distribution Learning.} We begin by learning position priors for the diffusion model across each large-scale dataset as shown in \cref{tab:exp:test_text}. Our results demonstrate that our approach outperforms other baselines in terms of RDA, achieving a high concentration of samples near the ground truth pose, with most poses being highly accurate. On smaller datasets, mixed training enhances performance. However, for larger datasets like Matrix City, mixed training leads to worse results, likely due to the increased ambiguity in text descriptions for large-scale scenes.

Additionally, we validated the effectiveness of the Gaussian splatting-based refinement method. By leveraging text CLIP embeddings and rendering image CLIP embeddings to filter and maximize similarity, we achieved significant improvements in distribution performance across most datasets.

\noindent\textbf{Granularity Experiment.}
We conduct experiments with varying levels of text granularity, categorizing them into (\textbf{Low Granulairty}, \textbf{High Granularity}, and \textbf{Max Granularity}). Note we define noun-based expressions as Low Granularity and sentence-based expressions as High Granularity as shown in \cref{sec:method}. For image inputs, we consider them as Max Granularity. As shown in \cref{tab:exp:test_text_granu_more}, our approach consistently outperforms the baseline, and the RDA score increases when more detailed text descriptions are provided, indicating that the estimated poses align more closely with the ground truth. Additionally, incorporating the most detailed image CLIP features yields the highest RDA score, as illustrated in \cref{fig:exp:qualitive_small_town_granu}. However, due to overlap and repetition in the city-level images, a substantial portion of the textual descriptions becomes duplicated. This duplication appears to diminish the benefit of mixing textual and image-based features, leading to our mixed approach being outperformed by the baseline (trained exclusively on images). A more detailed discussion of these observations is provided in \cref{suppsec:discuss}.

\subsection{Ablations \& User Studies}\label{sec:ablations}
\begin{figure}[ht]
    \centering
    \vspace{-1em}
    \includegraphics[width=0.95\linewidth]{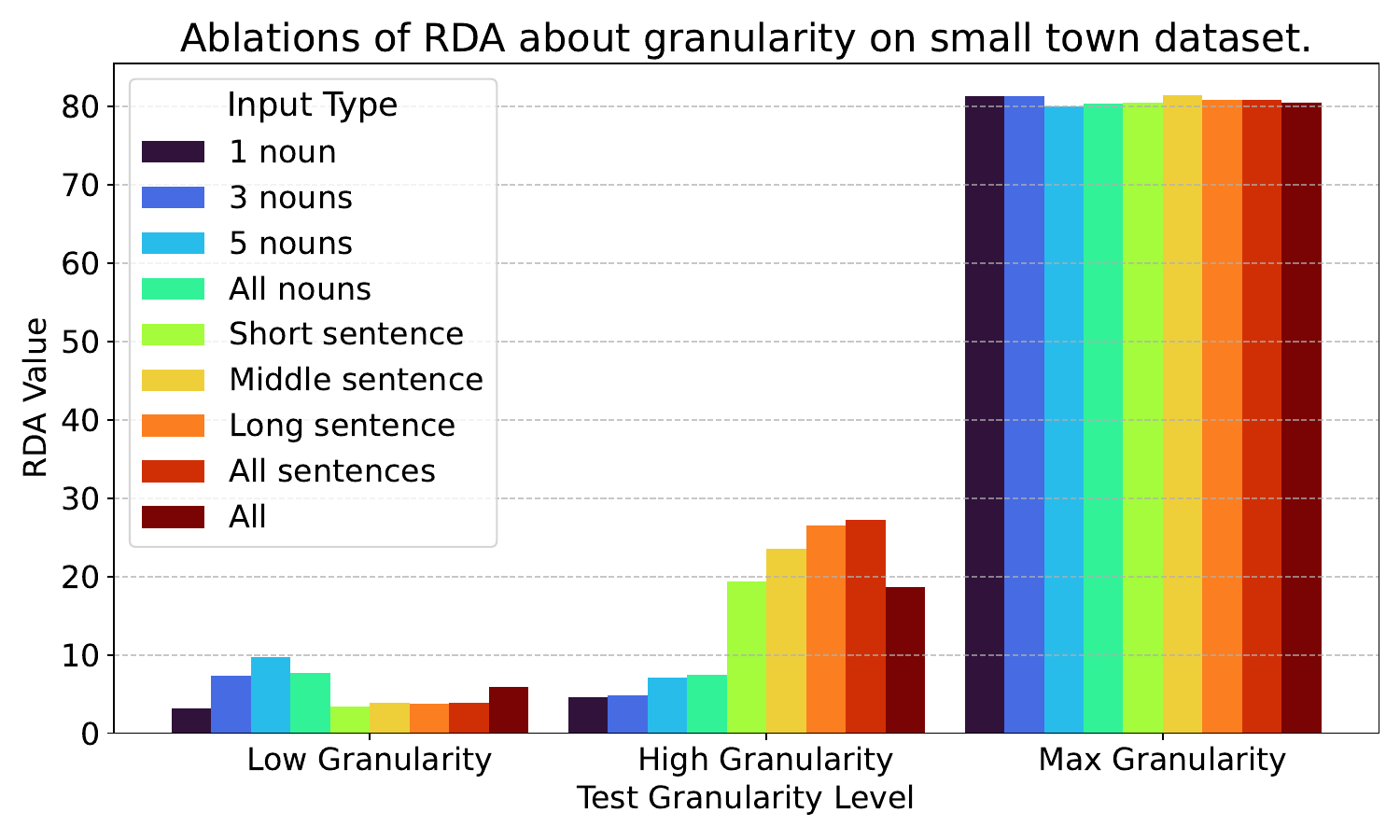}
    \caption{The bar chart compares RDA values across low, high, and maximum granularity levels for different input types for training, including varying numbers of nouns and sentences. The results indicate that increasing granularity and incorporating more descriptive text generally improve RDA performance.}
    \vspace{-1em}
    \label{fig:RDA-ablations}
\end{figure}

\noindent \textbf{Ablation on text granularity.} In \cref{fig:RDA-ablations}, we study how varying text granularity in training affects model performance on the small town dataset. Using the template from \cref{template}, we generate descriptions at four noun-based levels (\textbf{1 noun}, \textbf{3 nouns}, \textbf{5 nouns}, \textbf{All nouns}) and three sentence-based levels (\textbf{Short sentence}, \textbf{Middle sentence}, \textbf{Long sentence}). We further explore \textbf{All sentences} for every sentence length and \textbf{All} for every level of detail, ranging from single nouns to full paragraphs. This ablation provides a clear view of how each degree of textual detail influences overall model performance.

Results indicate that at low granularity, the \textbf{``All nouns''} setting achieves the highest RDA (7.68), suggesting that incorporating all nouns provides sufficient information for scene understanding in simple contexts. At high granularity, \textbf{``All sentences''} outperforms other settings with the highest RDA (27.27), indicating that detailed textual descriptions enhance the model's ability to capture complex spatial relationships. For max granularity, \textbf{``Middle sentence''} achieves the highest RDA (81.44), slightly outperforming \textbf{``1 noun''} (81.32), suggesting that moderate-length sentences effectively balance detail and generality. In terms of overall performance, the \textbf{``All''} setting provides a balanced approach, showing competitive RDA values across different granularity levels. Additionally, increasing sentence length from \textbf{``Short sentence''} to \textbf{``Long sentence''} improves RDA under high granularity, demonstrating that longer descriptions capture more scene details. Overall, finer text granularity using sentences performs better than using only nouns, with more detailed textual descriptions leading to improved performance under high granularity.

\begin{figure}[ht]
    \vspace{-2mm}
    \centering
    \includegraphics[width=0.8\linewidth]{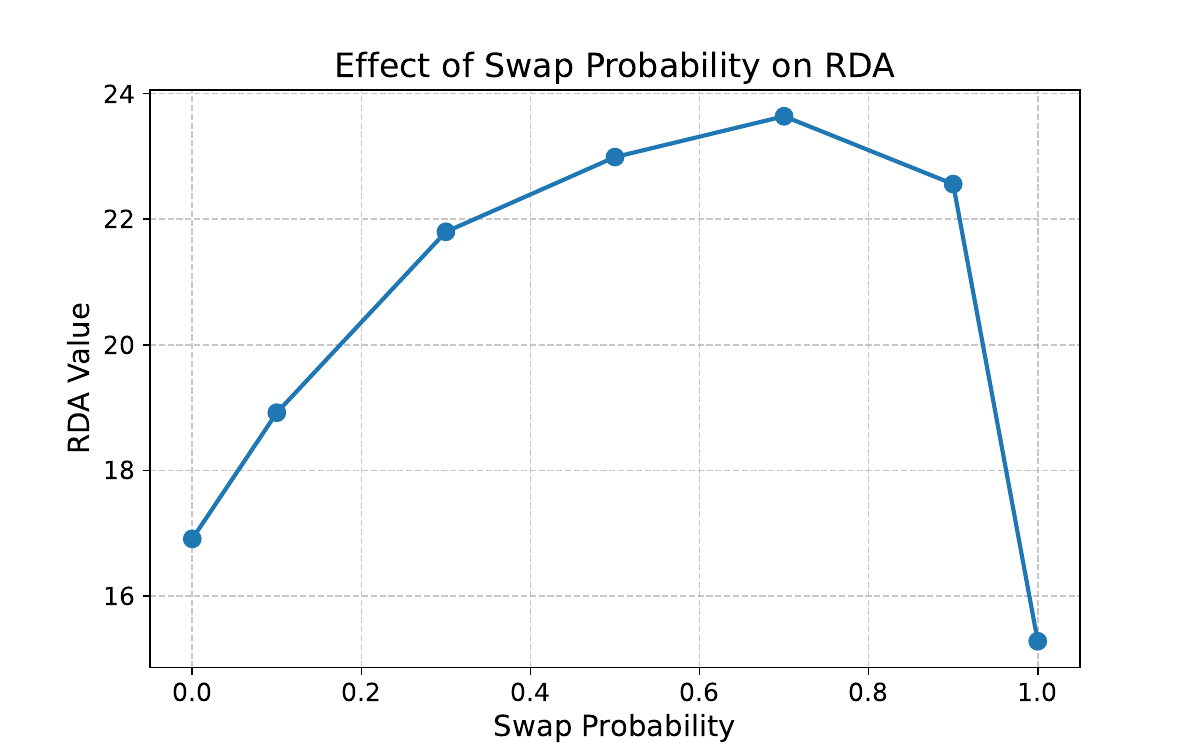}
    \vspace{-1mm}
    \caption{Ablations about swap probability.}
    \label{fig:RDA-swap-ablations}
    \vspace{-1em}
\end{figure}

\noindent \textbf{Ablation on swap ratio $\beta$}.
We conducted additional experiments on the swap ratio $\beta$ to investigate its effect on overall performance, as presented in \cref{fig:RDA-swap-ablations}. Results show that increasing the swap probability from $0$ to $0.7$ consistently boosts the RDA score, suggesting that introducing moderate diversity in the training data helps the model capture more robust representations. However, when the swap probability exceeds $0.7$, performance begins to degrade slightly, indicating that excessive swapping introduces too much noise. Overall, a moderate swap ratio of around $0.7$ appears to offer an optimal balance between enhanced diversity and noise control.

\noindent\textbf{User Studies.}
To validate the effectiveness of our method intuitively, we propose the user studies with question \textit{Which image do you think matches the text description?} (The answer can be multiple selections) as shown in \cref{fig:exp:visual_results}. 
\begin{wraptable}[4]{l}{40mm}
    \vspace{-4mm}
    \setlength{\extrarowheight}{2pt}
    \centering
    \renewcommand{\arraystretch}{0.9}
    \tabcolsep 2pt
    \resizebox{0.45\columnwidth}{!}{%
    \begin{tabular}{c|c|c}
        \toprule
        Poses	& GT  & Ours     	\\ 
        \midrule
        Match Rate (\%) & 56.5 & 50.8  \\ 
        \bottomrule
    \end{tabular}
    }
    \vspace{-3mm}
    \caption{User Study}
    \label{tab:exp:user_study}
\end{wraptable}
In particular, 10 samples are selected randomly from the output of our method (For each sample, there are 4 options from our method and 1 ground truth option). Then 30 volunteers are involved to answer this question. Then we compute the average match rate for GT and ours. From the results, in 30\% cases the performance of rendered images exceeds that of the ground-truth (GT) images. Overall, our rendered outputs are on par with the GT images, yielding performance comparable to real data. The corresponding results are shown in \cref{tab:exp:user_study}.

\section{Conclusion}
\label{sec:conclusion}
We presented a diffusion-based framework for text-based 6DoF localization at the city scale, addressing ambiguity and scalability challenges in large, complex environments. By leveraging 3D Gaussian splatting for efficient pose refinement through text and rendered image similarity optimization, our method achieves superior localization accuracy. Thanks to the Gaussian splatting-based scene representation that allows us to perform the visual reasoning during the refinement step. Applied across diverse large-scale scenes, our approach bridges text and visual understanding, enabling natural human-robot interaction and enhancing multi-modal reasoning in visual environments.



\clearpage
\bibliography{example_paper}
\bibliographystyle{icml2025}

\clearpage
\newpage
\appendix
In this appendix, we provide additional details to enhance the clarity and comprehensiveness of our work. First, we briefly illustrate two important base techniques, \ie Diffusion model and Gaussian Splatting in \cref{suppsec:priliminaries}. 
Next, the experimental setups are detailed in \cref{suppsec:exp_setup}, followed by an in-depth analysis of the ablation experiments in \cref{suppsec:ablation}, which demonstrate the effectiveness of our approach. 
Finally, we discuss further implications and insights in \cref{suppsec:discuss} and conclude with an exploration of the broader impact of our work in \cref{suppsec:impact}.

\section{Preliminaries} 
\label{suppsec:priliminaries}
\noindent{\textbf{Diffusion Models.}}
Diffusion models~\cite{ho2020denoising,sohl2015deep} are a class of likelihood-based generative models that approximate complex data distributions by inverting a diffusion process from data to a simple distribution via noising and denoising.
The noising process transforms data samples $x$ into noise over a sequence of $T \in \mathbb{N}$ steps. The model is trained to learn the denoising process.

A Denoising Diffusion Probabilistic Model (DDPM) specifies Gaussian noising.
For a variance schedule $\beta_1, ..., \beta_T$ across $T$ steps, the noising transitions are defined as follows:

\begin{equation}\label{eq:q}
q(x_t \mid x_{t-1}) = \mathcal{N}(x_t; \sqrt{1 - \beta_t} x_{t-1}, \beta_t \mathbb{I}),
\end{equation}
where $\mathbb{I}$ is the identity matrix.
This schedule ensures $x_T$ approaches an isotropic Gaussian, \ie, $q(x_T) \approx \mathcal{N}(\mathbf{0}, \mathbb{I})$.
Setting $\alpha_t = 1 - \beta_t$ and $\bar{\alpha}_t = \prod_{i=1}^t \alpha_i$ yields a closed-form solution for directly sampling $x_t$ given a data point $x_0$: 
\begin{equation}
x_t \sim q(x_t \mid x_0) = \mathcal{N}(x_t; \sqrt{\bar{\alpha}_t} x_0, (1 - \bar{\alpha}_t) \mathbb{I}).
\end{equation}

For sufficiently small $\beta_t$, the reverse $p_\theta(x_{t-1} \mid x_t)$ is Gaussian.
Thus, we approximate it with model $\mathcal{F}_\theta$:
\begin{equation}\label{eq:p}
p_\theta(x_{t-1} \mid x_t) = \mathcal{N}(x_{t-1}; \sqrt{\alpha_t} \mathcal{F}_\theta(x_t, t), (1 - \alpha_t) \mathbb{I}).
\end{equation}

\noindent{\textbf{3D Gaussian Splatting.}}
\label{subsec:gs_preliminary}
3DGS represents scene space with Gaussian primitives $\{Y_i\}_{i=1}^{N}$, stacking these as follows:
\begin{equation}
Y = [C, O, S, R, SH] \in \mathbb{R}^{N\times59},
\end{equation}
where $C \in \mathbb{R}^{N\times3}$ denotes the centroid, $O \in \mathbb{R}^{N\times1}$ the opacity, $S \in \mathbb{R}^{N\times3}$ the scale, $R \in \mathbb{R}^{N\times4}$ the quaternion rotation vector, and $SH \in \mathbb{R}^{N\times48}$ the spherical harmonics. These are collectively termed Gaussian parameters. Each Gaussian softly represents a spatial area with opacity. A point $q$ in the scene space is influenced by a Gaussian $Y_i$ according to the Gaussian distribution, weighted by opacity:
\begin{equation}
h_{i}(q) = O_i \exp\left( -\frac{1}{2} (q - C_{i})^T \Sigma_{i}^{-1} (q - C{i}) \right),
\end{equation}
where covariance $\Sigma_{i}$ is formulated as $\Sigma_{i} = R_i S_i S_i^{T} R_i^{T}$.

Projected onto a 2D image plane, each Gaussian’s influence, $h$, contributes to a pixel’s color through an alpha-blending equation over the set $\mathcal{G}$ of influencing Gaussians:
\begin{equation}
c_\text{pixel} = \sum_{i \in \mathcal{G}} c_i h^{\textrm{2D}}_{i} \prod_{j=1}^{i-1} (1 - h^{\textrm{2D}}_{j})\enspace.
\end{equation}

Through differentiable rasterization, rendering losses are back-propagated to update the Gaussian parameters. In this manner, we represent images rendered from pose $P$ given by $I=\mathcal{G}(P)$.



\begin{figure*}[t]
    \centering
    \includegraphics[width=0.85\linewidth]{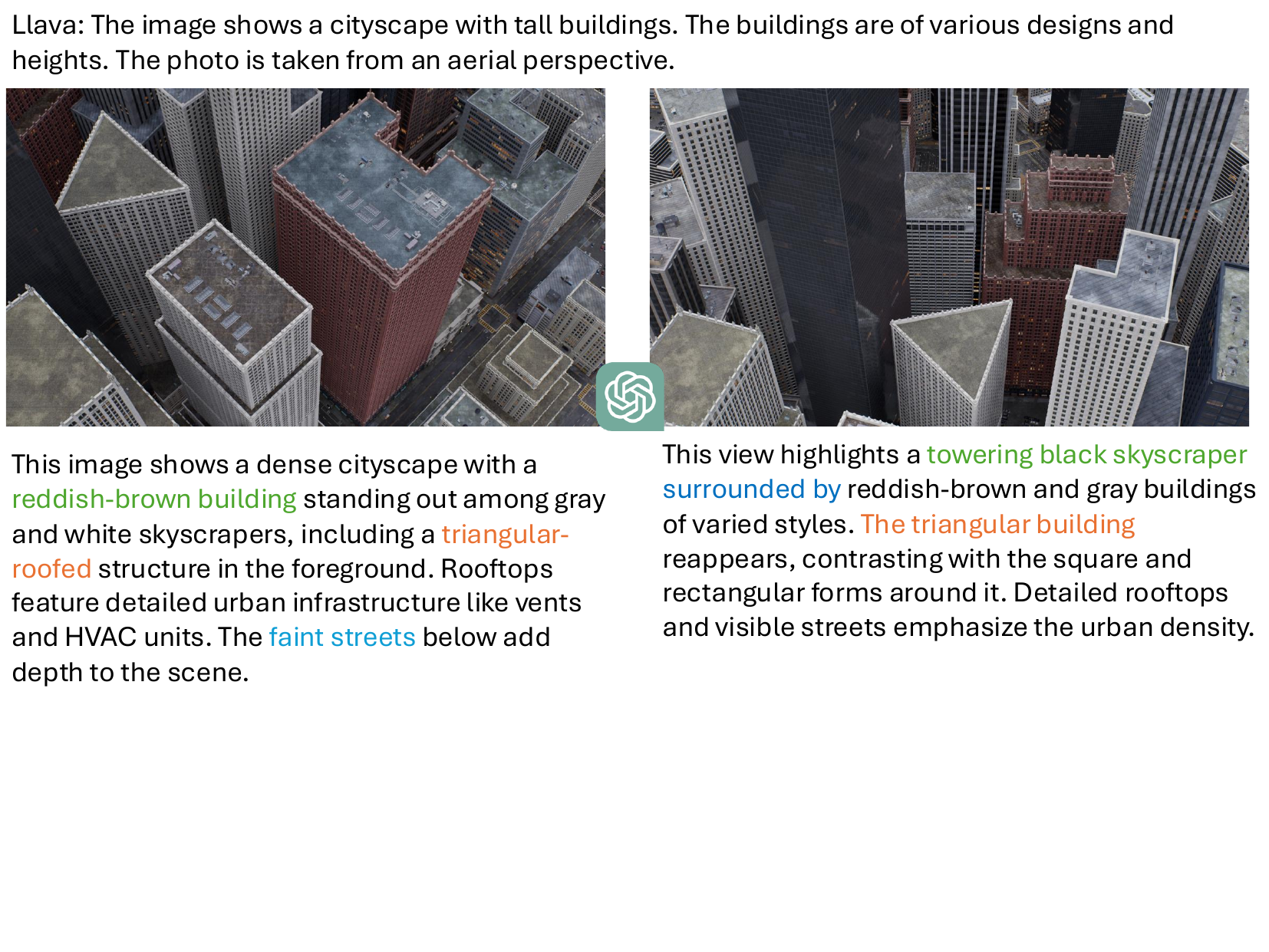}
    \vspace{-33mm}
    \caption{\textbf{Limitations on generated text for large scale scene:} We used the Llava model for image captioning at different granularities, but we found that for larger and more complex scenes, the text generated by Llava fails to extract more effective information. For example, as shown in the image above, for two pictures with numerous city buildings, Llava fails to generate distinct text prompts. In contrast, ChatGPT provides very detailed information that is helpful for localization.}
    \label{fig:exp:limitation}
    \vspace{-1em}
\end{figure*}

\section{Experimental Setups}
\label{suppsec:exp_setup}
\noindent \textbf{Gaussian Training}.
To process large-scale scenes, we utilize H3DGS~\cite{h3dgs} for Gaussian training. Chunk sizes are set to \(100 \times 100\) for street views and \(200 \times 200\) for aerial views. The datasets are partitioned into the following chunks: 4 for the small town dataset, 2 for the SciArt dataset, 4 for the residence dataset, 66 for the Matrix City aerial dataset, and 203 for the Matrix City small dataset. Each chunk is trained for 60,000 iterations. Subsequently, chunks are merged following hierarchy optimization; however, for the Matrix City dataset, memory constraints as shown in Fig.2 of our main manuscript prevent merging all chunks. Instead, only adjacent chunks are merged with overlap.
For each dataset, we first use the training views from regarding dataset to construct the Gaussian splats. Subsequently, we randomly select 10\% of the poses as the validation set, while the remaining 90\% are used to train the diffusion model. 

\noindent \textbf{Diffusion Model Training}.
In addition to the details mentioned in our main manuscript, for each batch, we randomly select 90 different timesteps out of a total of 100. We optimize the noise prediction step-by-step with a dropout rate of 0.1. The beta scheduler follows a linear progression from \(1 \times 10^{-4}\) to 0.1. For optimization, we use an \(L_1\) loss function for both translation and rotation quaternions. For the CLIP model, we use the pretrained laion2bs34b-b88k. For LLava, we utilize the lmms-lab/llama3-llava-next-8b model.

\noindent \textbf{Details of Gaussian based refinement}.
To identify the pose that best matches the text embedding at the feature level from the many possible outputs of a diffusion model, we propose the Gaussian Refinement Algorithm (\cref{alg:gaussian_refinement}). This algorithm iteratively refines an initial coarse pose by aligning its rendered view with the corresponding textual description in the embedding space, leveraging city-scale Gaussian models and the expressiveness of CLIP embeddings to achieve robust text-to-pose refinement.

The refinement process begins with a coarse camera pose, which is used to render a synthetic view based on the Gaussian model and pipeline parameters. During refinement, we follow the steps outlined in ICOMA, rendering images at a resolution of 224 $\times$ 224 while maintaining the same field of view (FoV). For images with differing aspect ratios, the shorter side is resized to 224 pixels, and the longer side is cropped to match the aspect ratio. Since the rasterization process only allows optimization of one image at a time, optimizing across all poses in the test set would be computationally prohibitive. Thus, refinement is conducted on a randomly selected 10\% subset of the test set.
In each refinement iteration, the rendered view is normalized and encoded using a CLIP Image Encoder, generating an embedding that is compared to the text embedding from a CLIP Text Encoder. A similarity score is computed between the two embeddings, and the loss is formulated as the negative of this similarity. Using gradient-based optimization, the coarse pose is iteratively updated to maximize the feature-level alignment with the text embedding.
To ensure robustness, the algorithm incorporates rejection mechanisms, discarding poses that fail to meet minimum similarity thresholds. 
By combining Gaussian models with CLIP embeddings and adhering to computationally efficient refinement steps, the \cref{alg:gaussian_refinement} effectively bridges the gap between text descriptions and 6DoF camera poses, delivering accurate localization results even in large-scale 3D environments.

\section{More Visual Results}
\label{suppsec:ablation}
\noindent \textbf{Qualitative results of granularity.}
We conducted experiments to evaluate the impact of text granularity across various datasets, and the results, illustrated in the Fig. 11 - 14, are consistent performs well. A pattern emerges across all datasets: as the text input becomes more detailed, the pose distribution transitions from a high-variance spread to a narrower, more precise distribution. In parallel, the estimated camera poses increasingly align with the given text input. Notably, when the most fine-grained input, such as an image, is used, our method achieves pose estimations that are the closest to the actual input pose.

\begin{figure*}[t]
    \centering
    \includegraphics[width=0.95\linewidth]{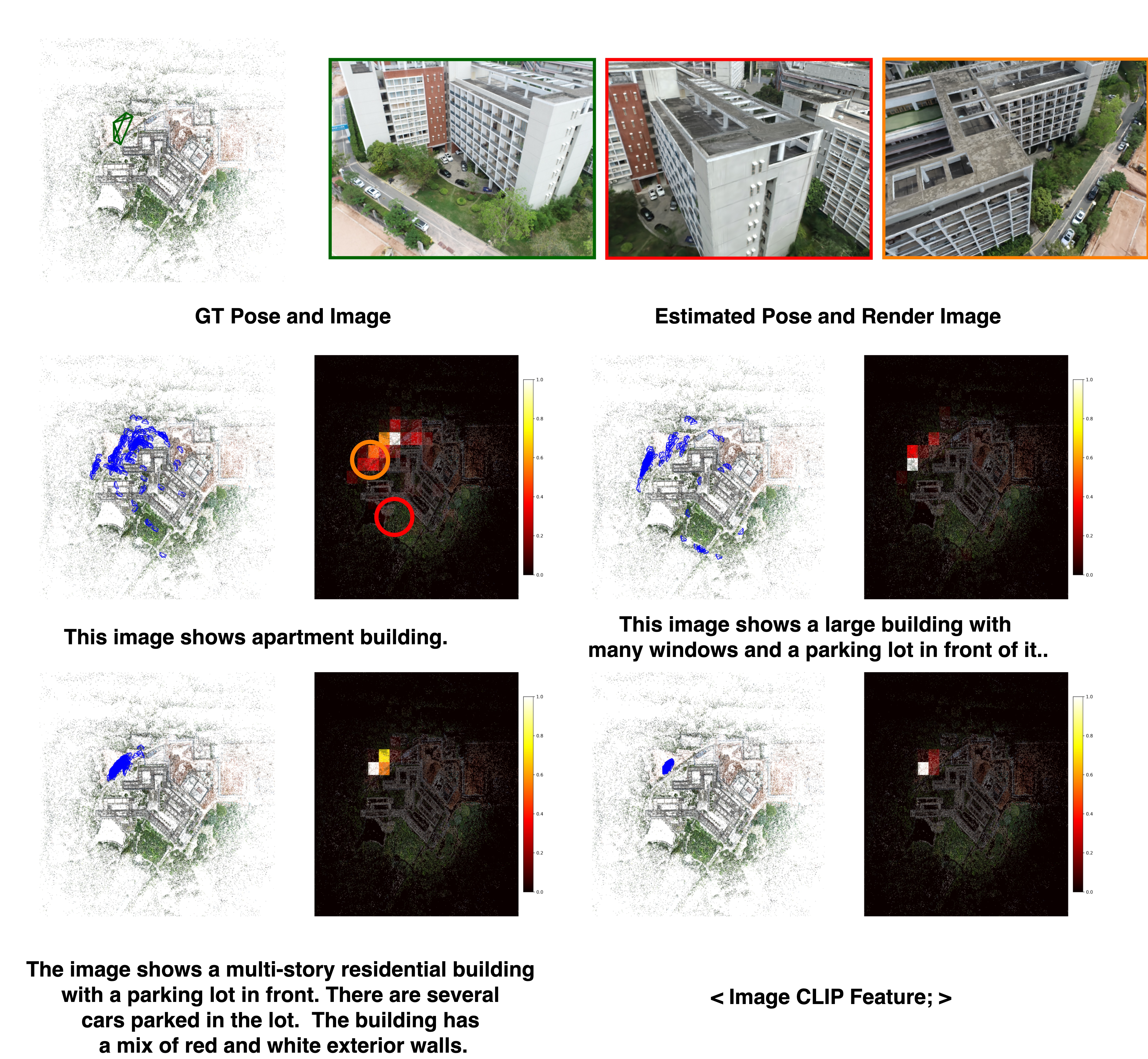}
    \vspace{-2mm}
    \caption{\textbf{Qualitative Results on the SciArt Dataset:} Similarly, we use the  \textcolor{green}{green camera} to indicate the camera pose used to generate the prompt, while high-density estimations are shown in \textcolor{taborange}{orange camera} and \textcolor{tabred}{red camera}. Providing more detailed text conditions results in a narrower distribution of estimated camera poses.}
    \label{fig:exp:qualitive_sciart_granu}
    \vspace{-1em}
\end{figure*}

\begin{figure*}[t]
    \centering
    \includegraphics[width=\linewidth]{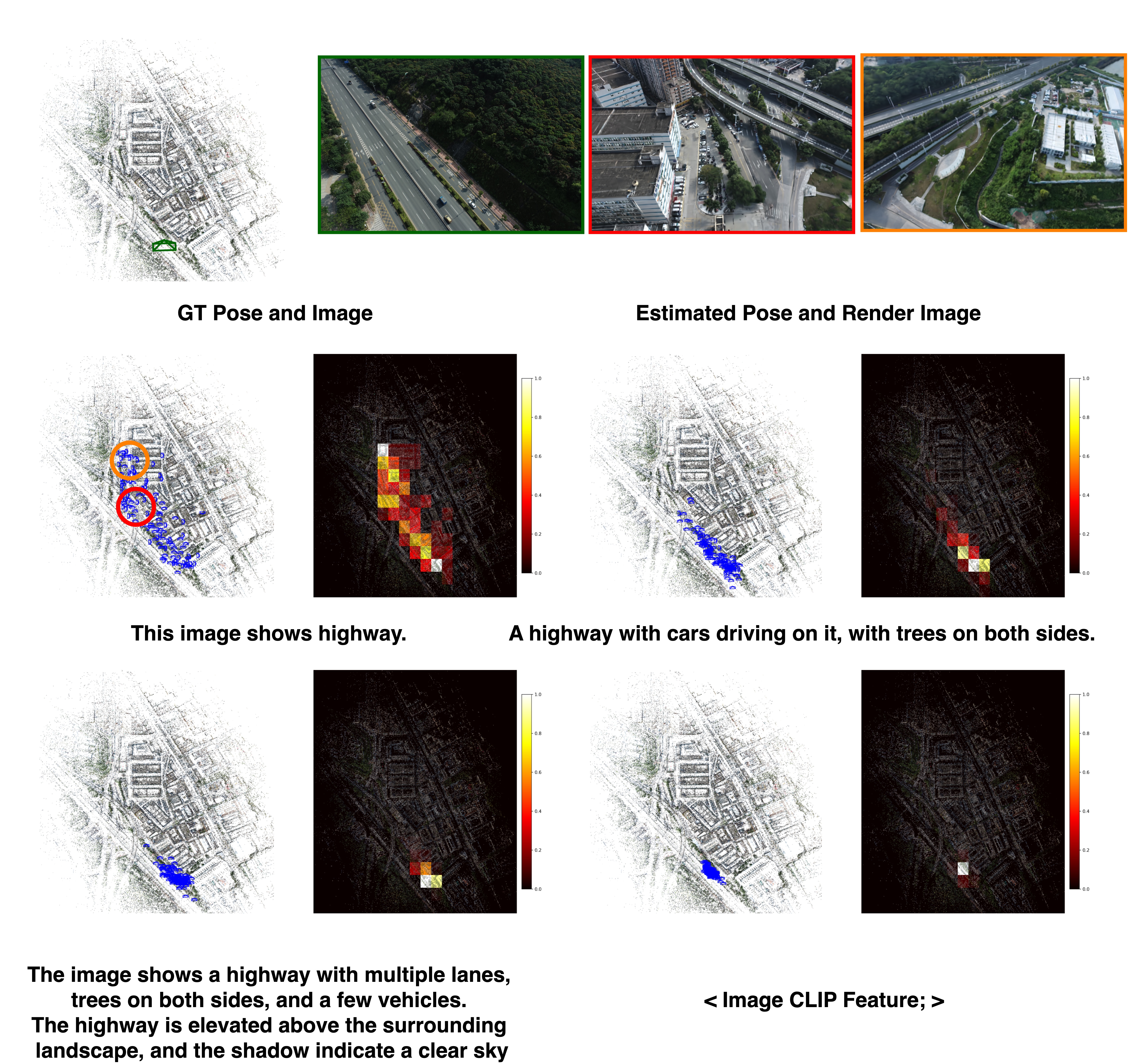}
    \vspace{-2mm}
    \caption{\textbf{Qualitative Results on the Residence Dataset:} The results demonstrate excellent localization potential. The dataset primarily features a residential area with a highway passing through the left section. When using a general prompt like "The image shows a highway," the estimated camera positions are distributed along the entire highway, successfully aligning with the text prompt. As more detailed information is provided, the estimated positions gradually narrow down and converge closer to the specific locations depicted in the images.}
    \label{fig:exp:qualitive_residence_granu}
    \vspace{-1em}
\end{figure*}

\begin{figure*}[t]
    \centering
    \includegraphics[width=0.95\linewidth]{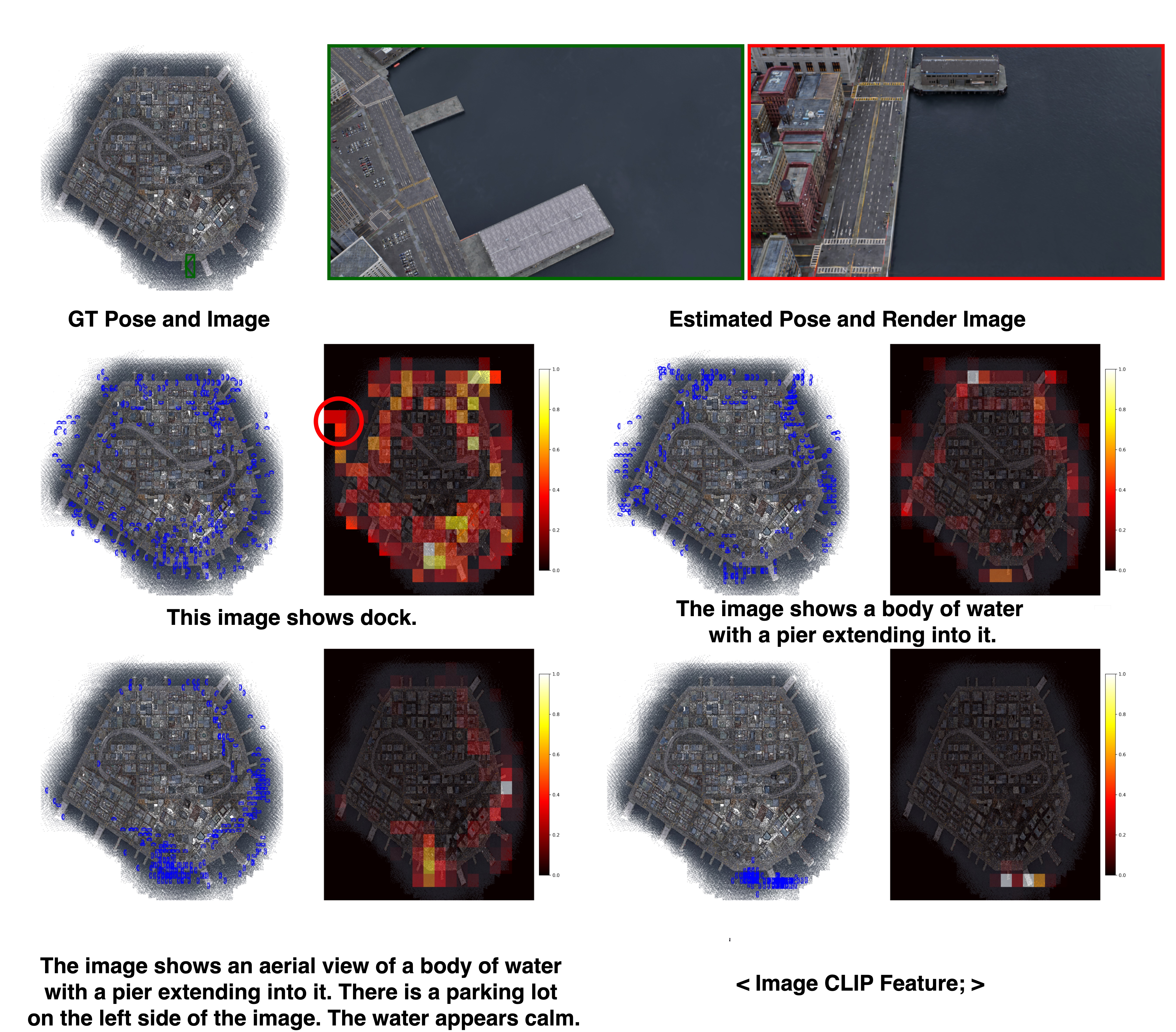}
    \vspace{-2mm}
    \caption{\textbf{Qualitative Results on the Matrix City Aerial Dataset:} We observe that our method also performs exceptionally well on large-scale scene datasets. For instance, the \textcolor{green}{green camera} highlights areas such as a dock and a parking lot. When using simpler or moderately detailed descriptions, the heatmap shows that the estimated poses are distributed across many similar scenes within the city that match the description. However, when the text input includes more specific details, such as mentioning the parking lot, the pose distribution suddenly narrows. The \textcolor{tabred}{red renderings} show areas without parking lots, where no poses are present in the estimated distribution.. Finally, when using the given image's CLIP features, we achieve the most accurate pose estimation.}
    \label{fig:exp:qualitive_aerial_granu}
    \vspace{-1em}
\end{figure*}

\begin{figure*}[t]
    \centering
    \includegraphics[width=\linewidth]{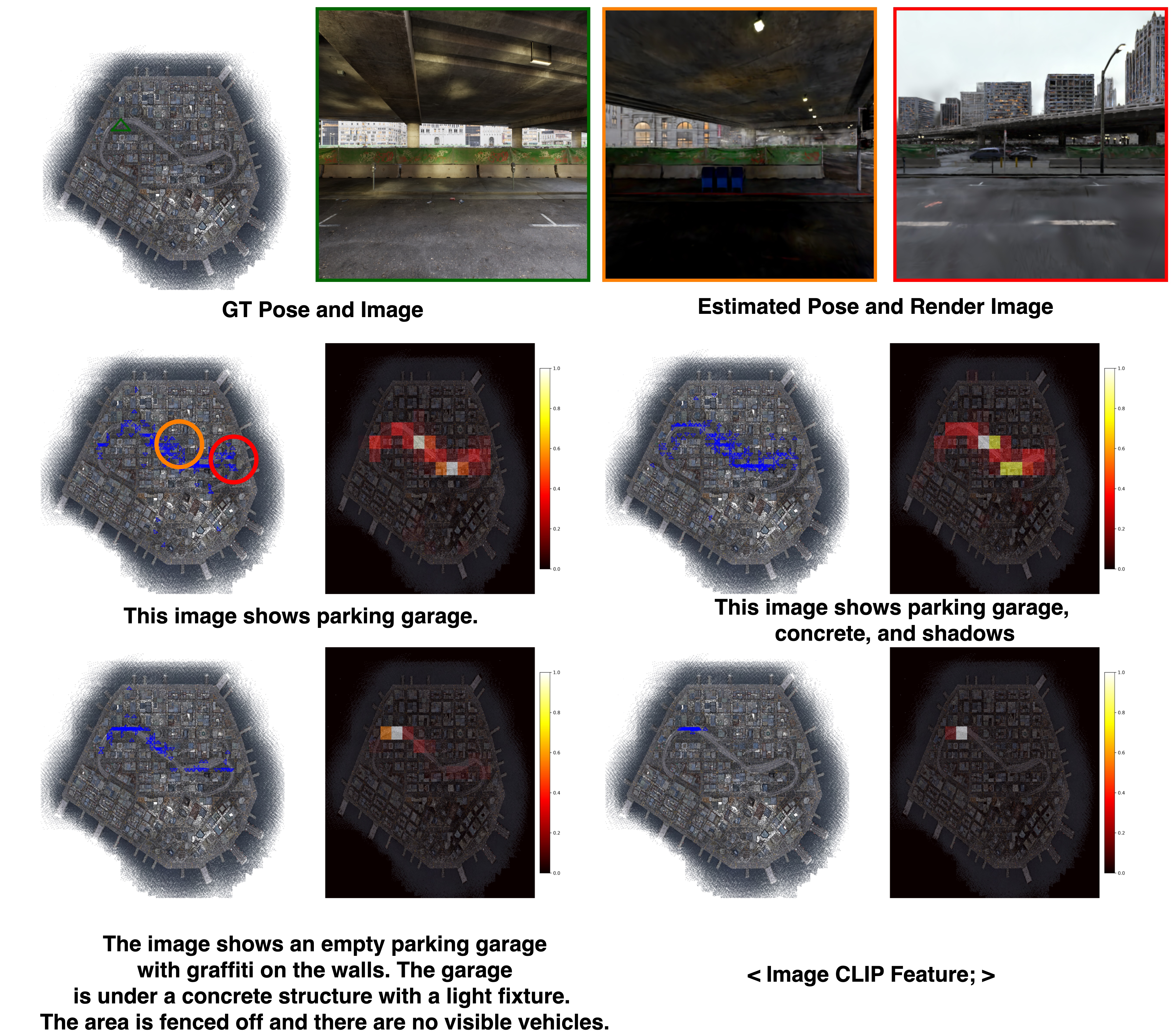}
    \vspace{-2mm}
    \caption{\textbf{Qualitative Results on the Matrix City Street Dataset:} On the urban street dataset, most street data is highly similar, with numerous overlapping text inputs. To address this, we used a more distinctive scenarioa garage under an overpass—as the image input. The results show that when we provide a coarse-grained input like "garage," our method generates poses near all potential parking locations, including not only under the overpass but also next to open-air street garages, as indicated by the \textcolor{tabred}{red renderings}. We also observe that as the input text becomes more specific, incorporating details like "shadow," "concrete," and "fence off," the distribution variance decreases, and the estimated poses converge closer to the input pose.}
    \label{fig:exp:qualitive_mc_street_granu}
    \vspace{-1em}
\end{figure*}

\begin{figure*}[!t]
    \centering
    \includegraphics[width=0.9\linewidth]{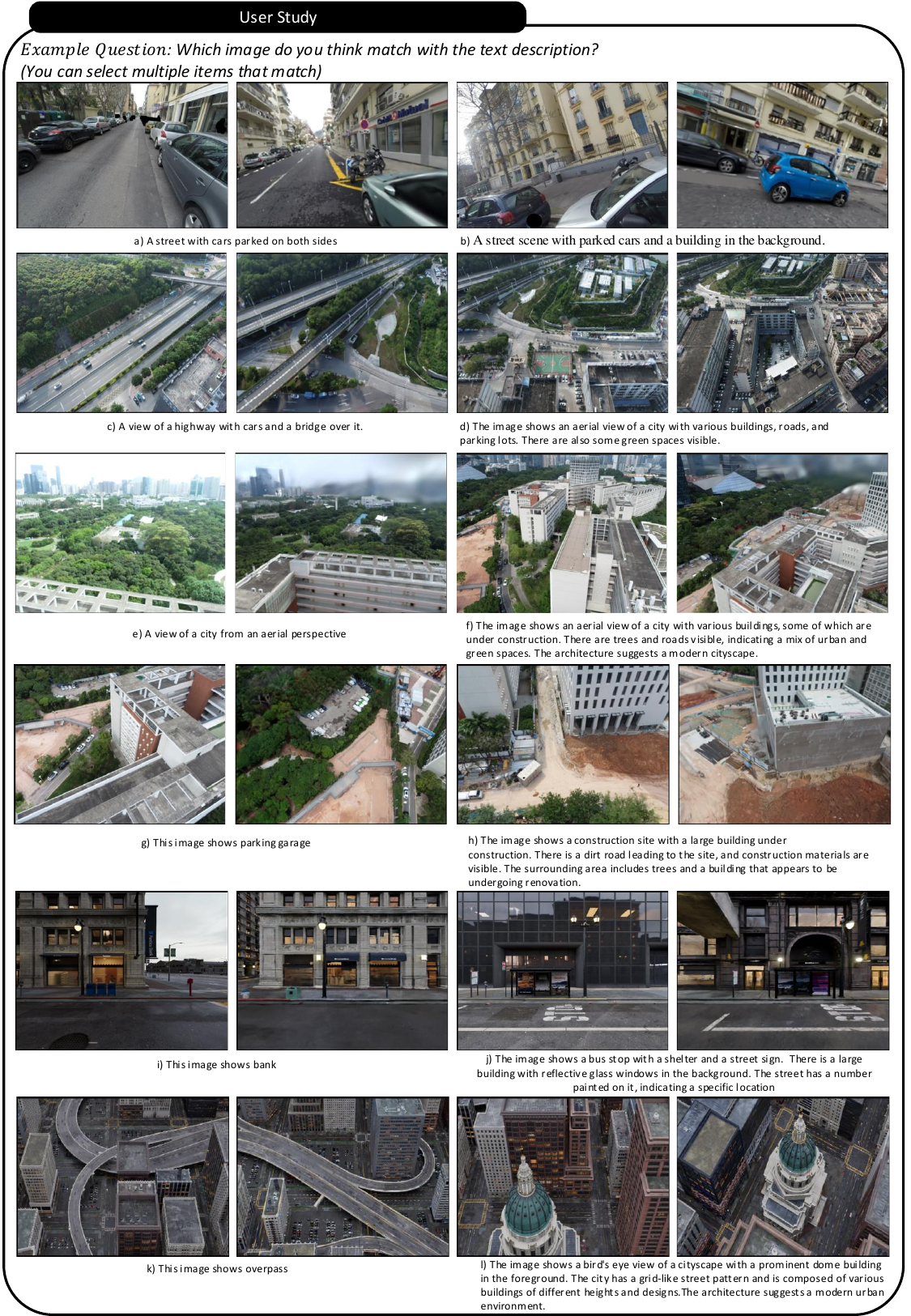}
    \vspace{-2mm}
    \caption{Qualitative results of the proposed method: Training images are shown on the left, with the rendered images on the right.}
    \label{fig:exp:user_study_more}
    \vspace{-1em}
\end{figure*}

\noindent \textbf{Qualitative Results of User Study.}
We present further qualitative results demonstrating how well the rendered images match the ground-truth images in response to various text descriptions, as depicted in \cref{fig:exp:user_study_more}. For instance, in image (a), the presence of cars parked on both sides of the street closely matches the described scenario, showcasing the effectiveness of our pose estimation technique. Image (d) particularly highlights the model’s capability to discern unique urban features, such as specific building types and road layouts, leading to precise pose estimations.

In images (b) and (c), the alignment of estimated camera poses with more generalized urban scenes shows the robustness of our approach across different settings. For example, in image (b), the estimation aligns well with a typical street scene, demonstrating the model's ability to contextualize and accurately generate relevant poses. 

Additionally, image (e) shows an aerial view of city parks juxtaposed against dense urban structures, a challenging scene that tests the limits of our model's contextual understanding. Despite this complexity, the model manages to provide a plausible estimate that aligns reasonably well with the text description. These examples underscore the potential of our method in practical applications, allowing for accurate and context-aware pose estimations based on varied textual inputs. There are more qualitative results in \cref{fig:exp:user_study_more}. 

\section{Discussion and Limitations}
\label{suppsec:discuss}
To the best of our knowledge, our work is the first to use a diffusion model based on text to estimate pose distributions. Compared to the baseline, our method effectively identifies scenes that match the text description. Additionally, the Gaussian refinement helps us filter out poor samples and further optimize the pose estimation. Our approach also supports multimodal image inputs.
In our experiments, we utilized the Llava model for image captioning at varying granularities. While the model performed well in many cases, we identified a significant limitation when applied to larger and more complex scenes. Specifically, Llava’s generated captions were unable to extract and convey sufficiently detailed information necessary for effective scene understanding, particularly in urban environments with dense structures as shown in \cref{fig:exp:limitation}.
Given these observations, we recognize the need for more powerful Visual Language Models (VLMs) that can generate richer and more accurate captions, especially for complex and large-scale scenes. As part of our future work, we plan to explore and incorporate more advanced VLMs to improve the quality and granularity of the text descriptions generated, which will ultimately enhance performance in tasks such as scene understanding, localization, and cross-modal retrieval.

\section{Impact Statement}
\label{suppsec:impact}
The proposed framework for text-based 6DoF camera localization introduces significant advancements in multi-modal reasoning, with potential impacts across various domains. By enabling precise localization based on ambiguous text descriptions, this method could transform applications such as autonomous navigation, urban planning, and augmented reality. For instance, in smart cities, this framework could aid in mapping and localizing critical infrastructure or guiding autonomous vehicles using natural language instructions.
Moreover, the integration of text and visual understanding enhances human-robot interaction, making it easier for non-experts to communicate with AI systems in large-scale, complex environments. The ability to perform visual reasoning with 3D Gaussian splatting also sets the stage for more efficient and scalable representations of urban and virtual scenes, benefiting areas like gaming, simulation, and virtual tourism.
However, the broad applicability of this technology may also raise ethical concerns, including potential misuse of surveillance or privacy violations. Ensuring the responsible deployment of such systems is essential to mitigate these risks. Overall, this work represents a step forward in bridging natural language understanding with spatial reasoning, paving the way for innovative multi-modal AI applications.


\end{document}